\documentclass{article}

\usepackage[utf8]{inputenc}
\usepackage[T1]{fontenc}
\usepackage[english]{babel} 	
\usepackage{hyperref}
\hypersetup{
  colorlinks   = true, 
  urlcolor     = blue, 
  linkcolor    = blue, 
  citecolor   = red 
}
\usepackage{apacite}
\usepackage{natbib}

\usepackage{xcolor}
\usepackage{tikz-qtree}
\usepackage{outlines}			
\usepackage{verbatimbox}
\usepackage{amsmath} 			
\usepackage{amssymb} 			
\usepackage{amsthm}				
\usepackage{stmaryrd} 			
\usepackage{braket}
\usepackage{listings}

\usepackage{marginnote}

\usepackage{geometry}
\DeclareMathOperator{\tran}{tran}
\DeclareMathOperator{\comp}{comp}

\newif\ifcomments
\commentstrue
\newcounter{note}
\stepcounter{note}

\theoremstyle{definition}
\newtheorem{definition}{Definition}

\lstdefinelanguage{Julia}%
  {morekeywords={abstract,break,case,catch,const,continue,do,else,elseif,%
      end,export,false,for,function,immutable,import,importall,if,in,%
      macro,module,otherwise,quote,return,switch,true,try,type,typealias,%
      using,while},%
   sensitive=true,%
   alsoother={$},%
   morecomment=[l]\#,%
   morecomment=[n]{\#=}{=\#},%
   morestring=[s]{"}{"},%
   morestring=[m]{'}{'},%
}[keywords,comments,strings]%

\author{Daniel Harasim (\'Ecole Polytechnique F\'ed\'erale de Lausanne)\\ Chris Bruno (McGill University)\\ Eva Portelance (Stanford University)\\ Martin Rohrmeier (\'Ecole Polytechnique F\'ed\'erale de Lausanne) \\ Timothy J. O'Donnell (McGill University)}
\title{A generalized parsing framework for Abstract Grammars\\Technical Report\thanks{Corresponding author: Daniel Harasim \texttt{daniel.harasim@epfl.ch}.}}
\date{\today}

\begin{document}

\maketitle
\section{Introduction}
\label{sec:intro}
This technical report presents a general framework for parsing a variety of grammar formalisms.
We develop a grammar formalism, called an Abstract Grammar, which is general enough to represent grammars at many levels of the hierarchy, including Context Free Grammars (CFGs), Minimalist Grammars (MG), and other weakly MG-equivalent languages like Linear Context-Free Rewriting Systems.
We then develop a single parsing framework which is capable of parsing grammars which are at least up to MGs on the hierarchy.
Our parsing framework exposes a grammar interface modelled on the Abstract Grammar formalism, so that it can parse any particular grammar formalism that can be reduced to an Abstract Grammar.

There is a great deal of previous work that is capable of parsing the grammars we treat here.
All of the grammars mentioned here have parsers written specifically for them and there are frameworks more general than the one given here, such as probabilistic programming languages that can specify arbitrary probabilistic programs.
Parsers for specific grammars are able to exploit optimizations specific to the task and formalism they were designed for and are therefore often faster than general systems.
However, these parsers have the disadvantage that they cannot be used for formalisms other than the one that they were intended for, making it difficult to prototype and compare different formalisms.
Our framework is the middle ground between these two approaches. We aim to be general enough to parse a variety of interesting formal grammars, while also exploiting optimizations specific to the parsing task.

In the following, we define Abstract Grammars as a generalization of Context Free Grammars where (i) the rewrite rules are partial functions and (ii) the set of nonterminals is paired with a set of operations to form a heterogeneous algebra.
We first define Abstract Context-free Grammars in \S\ref{sec:parsing:abstractcfg} which incorporate property (i), and then define the fully general Abstract Grammars in \S\ref{sec:parsing:mcsls} which incorporate property (ii).
By (i), generalizing the rewrite rules to any partial function, we can group related CFG rewrite rules into a common function, allowing those rules to share probability mass.
This is useful for representing a musical syntax where the the rules of prolongation and preparation, for example, are independent from the key \citep{rohrmeier2015towards, rohrmeier2011towards, lerdahl1985generative}.
By (ii) generalizing the nonterminals to elements of a heterogeneous algebra, we can represent languages higher than context-free on the hierarchy.
This is important for representing natural language, which occupies the space of mildly context-sensitive languages \citep{shiebe85, joshi85}.

In \S\ref{sec:parsing:abstractcfg}, after defining Abstract Context-free Grammars, we present a reduction automaton upon abstract CFGs, which is used to state an abstract grammar interface. We then describe the parsing algorithm, including code fragments of our Julia\footnote{\href{https://julialang.org}{julialang.org}, see also \cite{bezanson2017julia}} implemetation.
\S\ref{sec:parsing:mcsls} defines Abstract Grammars and revises the reduction automaton and the grammar interface given in \S\ref{sec:parsing:abstractcfg} for the fully general case.
In \S\ref{sec:mgs}, we show how to state specific interface functions for a Minimalist Grammar, and an implementation of this interface in Julia.

\section{Parsing Abstract Context-free Grammars}
\label{sec:parsing:abstractcfg}
\subsection{Abstract Context-free Grammars}
\label{sec:abstractcfg}

\begin{definition}
A \textit{(deterministic) Abstract Context-free Grammar} $G = (T,N,S,\Gamma)$ consists of a set $T$ of \textit{terminal categories}, a finite set $N$ of \textit{non-terminal categories}, a set of partial functions $$\Gamma := \set{g | g:N \nrightarrow (T\cup N)^*},$$ called \textit{rewrite functions} or \textit{generation functions}, and a \textit{start category} $S\in N$. We denote the set $T\cup N$ of all categories by $C$. A sequence of categories $\beta\in C^*$ can be \textit{generated in one step} from a sequence $\alpha\in C^*$ by the application of a rewrite function $g\in \Gamma$ to a non-terminal category $A\in N$, denoted by $\alpha\longrightarrow_{g(A)} \beta$, if $\alpha=\alpha_1 A \alpha_2$ and $\beta=\alpha_1 g(A) \alpha_2$ for some $\alpha_1,\alpha_2\in C^*$. We write $\alpha\longrightarrow\beta$ if there exists any $g\in\Gamma$ and $A\in N$ such that $\alpha\longrightarrow_{g(A)}\beta$. The transitive closure of the generation-in-one-step relation $\longrightarrow$ is denoted by $\longrightarrow^*$. The language of the grammar $G$ is the set of sequences of terminal categories that can be generated from the start category $S$, that is $\mathcal L(G) = \set{\alpha\in T^* | S\longrightarrow^*\alpha}$.
\end{definition}

The languages that can be described by Abstract Context-free Grammars are exactly the languages that can be described by classical Context-free Grammars. For each classical Context Free Grammar $(T,N,S,R)$ where $R\subseteq N\times C^*$, we can construct an Abstract Context-free Grammar $(T,N,S,\Gamma)$ which generates the same language and vice versa. By setting $\Gamma:=\set{N\nrightarrow C^*, A\mapsto \alpha | (A,\alpha)\in R}$ (considering each rewrite rule as a single partial function), each classical Context-free Grammar induces an equivalent Abstract Context-free Grammar. In converse, each Abstract Context-free Grammar induces an equivalent classical Context-free Grammar by setting $R:=\bigcup_{g\in \Gamma} \set{(A,\alpha)\in N\times C^* | g(A)=\alpha}$, because the set of non-terminal categories $N$ is finite. Therefore, Abstract Context-free Grammars are essentially classical Context-free Grammars where rules are glued together respecting the well-definedness of (partial) functions. They do not add deterministic expressiveness, but enable one to assign probabilities to sets of rules. Since Abstract Context-free Grammars use partial rewrite functions instead of classical rewrite rules, they moreover do not treat non-terminal categories as atomic symbols like classical Context-free Grammars, but enable non-terminal categories of any data type.

\begin{definition}
A \textit{probabilistic Abstract Context-free Grammar} is a deterministic Abstract Context-free Grammar where each non-terminal category $A\in N$ is associated with a random variable $X_A$ over rewrite-functions.
The probability of a sequence $\alpha\in C^*$ rewriting in one step into a sequence $\beta\in C^*$ by applying a rewrite function $g\in\Gamma$ to a non-terminal $A \in N$ is
\[
	P(\alpha \longrightarrow_{g(A)} \beta) = \text{splits}(g,A,\alpha,\beta) \cdot P(X_A = g),
\]
where
$$\text{splits}(g,A,\alpha,\beta) = \#\set{(\alpha_1,A,\alpha_2)\in C^*\times N \times C^* | \alpha_1 A \alpha_2 = \alpha \text{ and } \alpha_1 g(A) \alpha_2 = \beta}$$
denotes the number of occurrences of the non-terminal category $A$ in the sequence $\alpha$ to which $g$ can be applied to get $\beta$. 
The probability of any sequence $\alpha$ rewriting into $\beta$ in one step is then
\[
	P(\alpha \longrightarrow \beta) = \sum_{g \in \Gamma} \sum_{A \in N} P(\alpha \longrightarrow_{g(A)} \beta).
\]
The probability of a sequence $\alpha$ rewriting into a sequence $\beta$ in any steps is recursively defined as
$$P(\alpha\longrightarrow^* \beta) = \sum_{\alpha'\in C^*} P(\alpha\longrightarrow^* \alpha')P(\alpha'\longrightarrow \beta).$$
The probability that a sequence of categories $\alpha$ generated by the grammar $\mathcal G$ is the probability that $\alpha$ is generated from the grammar's start symbol, that is $P(\alpha) = P(S\longrightarrow^*\alpha)$.
\end{definition}

Probabilistic Abstract Context-free Grammars are more expressive than classical probabilistic Context-free Grammars, in the sense that they can express a wider range of probability distributions over rules. In a probabilistic Abstract Context-free Grammar, non-terminal categories can share the same probability distribution over rewrite functions without rewriting to exactly the same right-hand sites. Consider for example a probabilistic Context-free Grammar that contains the characters a, b, A, and B as non-terminal categories, a rewrite function $g$ that capitalizes non-capitalized characters ($g(\text{a})=\text{A}$ and $g(\text{b})=\text{B}$), and a rewrite function $h$ that swaps as and bs ($g(\text{b})=\text{a}$ and $g(\text{a})=\text{b}$). By putting probability mass on $g$ in the distributions associated with a and b, this grammar could for example then learn easily from data that the abstract concept of capitalization is more probable than swapping. In contrast, a classical probabilistic Context-free Grammar would have to learn that capitalizing is more probable than swapping separately for each non-terminal symbol. We hope that this feature is helpful for processing musical data in which the usage of a rewrite function is assumed to be independent from the key of the musical objects its applied to.

\subsection{Reduction automata and the abstract grammar interface}
\label{sec:automaton}
\begin{figure}
\label{search trie}
\centering
\includegraphics[width=0.9\textwidth]{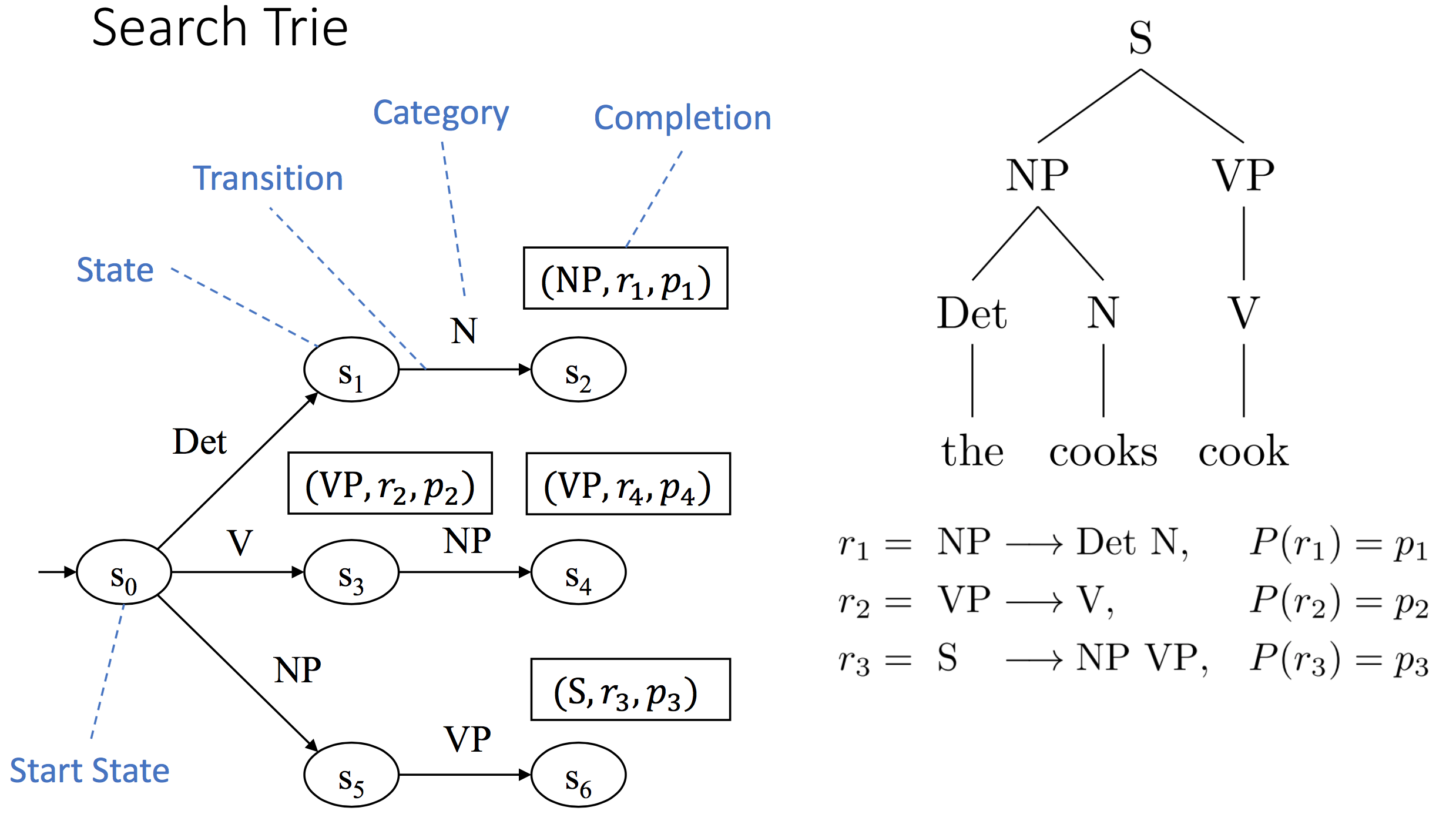}
\caption{Illustration of the reduction automaton construction for Abstract Context-free Grammars with a bottom-up parsing strategy. Only the rules that rewrite non-terminal categories to sequences of non-terminal categories are shown.}
\end{figure}

This section shows how to construct an automaton that accepts exactly the sequences that are generatable by a deterministic Abstract Context-free Grammar using the notions of states $\mathcal S$, a transition function $\text{tran}:\mathcal S\times (T\cup N)^* \to\mathcal S$, and a completion function $\text{comp}:\mathcal S\to 2^N$, where $2^N$ denotes the powerset of the set of non-terminal categories. A definition of this automaton corresponds to a parsing strategy such as Earley parsing or CYK-like bottom-up parsing. In general, however, the transition function on the set of states and category sequences can be any finite-state automaton. In the following, we construct a bottom-up parsing automaton that accepts the language of a given functional Context-free Grammar. The construction is illustrated in Figure \ref{search trie}. The set of states is simply the set of category sequences, $\mathcal S=(T\cup N)^*$, the transition function is the concatenation of category sequences and the completion function is the union of all preimages under all rewrite functions, $\text{comp}(s) = \bigcup_{g\in \Gamma}g^{-1}(s)$. Therefore,
\begin{align*}
\alpha A\gamma \longrightarrow \alpha\beta\gamma
	&\quad\iff\quad A \longrightarrow \beta \\
	&\quad\iff\quad \exists g\in \Gamma: g(A)=\beta \\
	&\quad\iff\quad A\in \bigcup_{g\in \Gamma}g^{-1}(\beta) \\
	&\quad\iff\quad A\in \text{comp}(\text{tran}(\varepsilon, \beta))
\end{align*}
for all non-terminal categories $A\in N$ and sequences $\alpha,\beta,\gamma\in (T\cup N)^*$, where $\varepsilon$ denotes the empty sequence. We thus can abstract from our concrete construction and define an abstract context-free reduction automaton.

\begin{definition}
A \textit{reduction automaton} $\mathcal{A}=(T,N,S,\mathcal S,s_0,\tran,\comp)$ consists of a set of \textit{terminal categories} $T$, a set of \textit{non-terminal categories} $N$, a \textit{goal category} $S\in N$, a set of \textit{states} $\mathcal S$, an \textit{initial state} $s_0 \in \mathcal S$, a \textit{transition function} $\tran: \mathcal S \times (T\cup N)^* \to \mathcal S$, and a \textit{completion function} $\comp:\mathcal S \to 2^{N \times \Gamma}$.

For $\alpha,\beta,\gamma\in (T\cup N)^+$ and $A \in N$, the sequence $\alpha A \gamma$ is called a \textit{reduction} of $\alpha\beta\gamma$ iff $(A, g)$ is a completion of the state $\tran(s_0,\beta)$ for some rewrite function $g$,
$$\alpha A \gamma \longrightarrow \alpha\beta\gamma \quad :\iff \quad \exists g \in \Gamma: (A, g) \in \comp(\tran(s_0,\beta)).$$
The automaton accepts a sequence of terminal categories $\alpha$ if it can be reduced to the goal category $S$. The language $\mathcal{L}(\mathcal{A})$ of the automaton $\mathcal{A}$ is thus defined as all accepted sequences. That is,
$$\alpha \in \mathcal{L}(\mathcal{A}) \quad :\iff \quad S \longrightarrow^* \alpha,$$
for all $\alpha\in T^*$, where $\longrightarrow^*$ denotes the transitive closure of $\longrightarrow$.
\end{definition}

\paragraph{Abtract grammar interface} Remark that to gain computational efficiency, the transition function can rigorously filter its values so that it only yields states that eventually can lead to completions.
For a probabilistic Abstract Context-free Grammar, the completion function is modified to $\text{comp}:\mathcal S\to 2^{N \times \Gamma \times [0,1]}$. In our construction,
$$\text{comp}(s)=\bigcup_{g\in \Gamma}\set{(A,p)\in N\times[0,1] | g(A)=s, p=P(X_A = g)}.$$ 
We use this reduction automaton to state the abstract grammar interface in figure \ref{interface}. The power of this interface is that it can be used even for mildly context-sensitive grammars as we show in later sections.

\begin{figure}[!htb]
\label{interface}
\begin{alignat*}{3}
&\text{tran\_possible} &&\colon \mathcal S \times (N\cup T)^* &&\rightarrow \set{\text{True},\text{False}} \\
&\text{tran} &&\colon \mathcal S \times (N\cup T)^* &&\nrightarrow \mathcal S \\
&\text{comp} &&\colon \mathcal S &&\rightarrow 2^{N\times \Gamma \times[0,1]} \\
&\text{startstates} &&\colon ~ &&\rightarrow 2^\mathcal S \\
&\text{startcategories} &&\colon ~ &&\rightarrow 2^N
\end{alignat*}
\caption{The abstract grammar interface. $\mathcal S$ denotes the set of states, $T$ denotes the set of terminal categories, $N$ denotes the set of non-terminal categories, $2^N$ denotes the powerset of $N$, etc. Since the transition function tran is a partial function, a function tran\_possible is used to test whether a transition is defined for a given pair of a state and a category. The functions startstates and startcategories are constant functions and thus denote without a domain.}
\end{figure}

\paragraph{Parsing as deduction rules}
Figure \ref{fig:cfg.deduction} presents a generic parsing algorithm for the abstract grammar interface using the parsing as deduction framework \citep{Shieber1993, Goodman1998}. The main difference between the parser and the grammar is that the parser is able to access the indices of the terminal categories of a given input sequence, while this information is not accessed by the grammar itself. Given a context-free reduction automaton $\mathcal{A}=(T,N,S,\mathcal S,s_0,\tran,\comp)$, we call a state $s\in\mathcal S$ together with a start index $i$ and an end index $j$ an \textit{edge} and denote it by $[s,i,j]$. Analogously, we call a non-terminal category $A\in N$ together with start and end indices $i$ and $j$ a \textit{constituent} and denote it by $[A,i,j]$. Figure \ref{fig:cfg.deduction} shows the goal item and the deduction rules.

\begin{figure}[!htb]
\begin{alignat*}{3}
&\text{items:} && \text{edges}\qquad\qquad[s,i,j] \quad\,\text{ for } s\in \mathcal S \text{ and } i,j\in\{1,\dots,|w|+1\} \\
& &&\text{constituents}\quad[A,i,j] \quad\text{ for } A\in N \text{ and } i,j\in\{1,\dots,|w|+1\} \\ \\
&\text{goal item:} && [S,1,|w|+1] \\ \\
&\text{axioms:} && \frac{}{[w_i,i,i+1]}  \quad\text{ for } i\in\{1,\dots,|w|\} \\ \\
&\text{introduce edge:} && \frac{[A,i,j]}{[s,i,j]} \quad s=\text{tran}(s_0, A) \\ \\
&\text{complete edge:} && \frac{[s,i,j]}{[A,i,j]} \quad \exists g \in \Gamma: (A, g) \in\text{comp}(s) \\ \\
&\text{fundamental rule:}\qquad\qquad && \frac{[s,i,j] \quad [A,j,k]}{[s',i,k]} \quad \text{tran}(s,A) = s'
\end{alignat*}
\caption{Description of the parsing algorithm by the use of deduction rules}
\label{fig:cfg.deduction}
\end{figure}

\subsection{Parser implementation and data type description}
\label{sec:implementation}

\begin{figure}
  \centering
\begin{lstlisting}
function run_chartparser(grammar, input)
    chart, agenda, logbook = initialize(grammar, input)
    while !isempty(agenda)
        item = dequeue!(agenda)
        finish!(item, chart, agenda, logbook, grammar)
    end
    return ParseForest(chart, logbook, input, grammar)
end

function finish!(item, chart, agenda, logbook, grammar)
    if no_noteworthy_inside_score_change_since_its_last_dequeue(item)
        insert!(chart, item)
        do_fundamental_rule!(item, chart, agenda, logbook, grammar)
    else
        inference_rule(item)(item, agenda, logbook, grammar)
        update_inside_score_tracker!(item)
        re_enqueue!(agenda, item)
    end
end

inference_rule(edge::Edge) = complete_edge!
inference_rule(cons::Constituent) = introduce_edge!
\end{lstlisting}
  \caption{A simplified code-fragment of the parser implementation}
  \label{fig.parsing algorithm}
\end{figure}

The Julia implementation of the parsing algorithm is shown in Figure \ref{fig.parsing algorithm}. It is manly inspired by \cite{klein2004parsing}. It implements the above presented parsing as deduction rules using the data types \texttt{Item}, \texttt{Edge}, \texttt{Constituent}, \texttt{ItemKey}, \texttt{EdgeKey}, \texttt{ConstituentKey}, \texttt{Traversal}, \texttt{Completion}, \texttt{Agenda}, \texttt{Chart}, and \texttt{Logbook} that we explain in the following. Edges and constituents as shown in the parsing as deduction framework have the type \texttt{Edgekey} and \texttt{ConstituentKey}, respectively. \texttt{ItemKey} is essentially the type union of \texttt{Edgekey} and \texttt{ConstituentKey}.
\newpage
\begin{lstlisting}
immutable EdgeKey{St} <: ItemKey
    state :: St
    start :: Int
    end_  :: Int
end

immutable ConstituentKey{C} <: ItemKey
    cat   :: C
    start :: Int
    end_  :: Int
end
\end{lstlisting}
\vspace{10px}

Instances of the types \texttt{Edge} and \texttt{Constituent} are edge keys and constituent keys equipped with additional information such as their inside scores, back-pointers to the items from which they were created from, and unique IDs that are used as pointers. For better readability, we only present additional information here that is crucial to the parsing algorithm. The type \texttt{Item} is again essentially the type union of \texttt{Edge} and \texttt{Constituent}.

\vspace{10px}
\begin{lstlisting}
type Edge{St,S} <: Item
    key         :: EdgeKey{St}
    score       :: S
    traversals  :: Vector{Traversal{S}}
    id          :: Int
    lastpopprob :: LogProb
end
type Constituent{C,S} <: Item
    key         :: ConstituentKey{C}
    score       :: S
    completions :: Vector{Completion{S}}
    id          :: Int
    lastpopprob :: LogProb
end
\end{lstlisting}

\newpage
The types \texttt{Traversal} and \texttt{Completion} are the back pointers of edges and constituents, respectively. Every traversal describes exactly one way of building an edge and every completion describes exactly one way of building a constituent. Since there can be multiple ways to build edges and constituents, edges contain a list of traversals and constituents contain a list of completions from which they were created or updated. We additionally distinguish between edge completions and terminal completions and implement the \texttt{Completion} type as an abstract data type.

\vspace{10px}
\begin{lstlisting}
immutable Traversal{S}
    edgeid  :: Int
    consid  :: Int
    score   :: S
end
immutable EdgeCompletion{S} <: Completion{S}
    edgeid :: Int
    score  :: S
end
immutable TerminalCompletion{T,S} <: Completion{S}
    terminal :: T
    score    :: S
end
\end{lstlisting}
\vspace{10px}

The \texttt{Logbook} is a hash table that maps item keys to their items. It is in particular used after an inference rule was applied to point to the inferred item if it has been created already.

\vspace{10px}
\begin{lstlisting}
type ParserLogbook{C,St,S}
    edges   :: Vector{Edge{St,S}}
    conss   :: Vector{Constituent{C,S}}
    edgeids :: Dict{EdgeKey{St}, Int}
    consids :: Dict{ConsKey{C}, Int}
end
\end{lstlisting}
\vspace{10px}

Each item is at any point of the parsing process either stored in the agenda or the chart. The agenda (singleton of type \texttt{Agenda}) is a priority queue that contains the items for which the ways to construct them are not all known yet. Items of small length are favored and edges are preferred over constituents to implement a bottom-up chart parsing. In contrast, the chart (singleton of type \texttt{Chart}) stores the items of which all ways to create them are known. Edges are stored by their stored by their end index and constituents by their start index to speed up the processing of the fundamental rule.

\vspace{10px}
\begin{lstlisting}
type ChartCell{C,St}
    edgeids :: Dict{St, Vector{Int}}
    consids :: Dict{C, Vector{Int}}
end

type Chart{C,St}
    cells :: Vector{ChartCell{C,St}}
end
\end{lstlisting}

\section{Parsing mildly-context sensitive languages with Abstract Grammars}
\label{sec:parsing:mcsls}
\subsection{Abstract Grammars}
\label{sec:ags}
This section generalizes Abstract Context-free Grammars to capture mildly context-sensitive structures.
Recall that in the context-free case, the rewrite arrow $\longrightarrow^*$ is a binary relation between sequences of terminal and non-terminal categories. If we eventually end up with a sequence of terminal categories in our generation process, we simply concatenate the terminal symbols.
In an Abstract Grammar, we generalize that concatenation to other algebraic operations stated upon \emph{tuples} of terminal symbols.
To do this, we define \emph{heterogeneous algebras} and use them to define an Abstract Grammar.
In the following, it might be helpful to keep in mind the word monoid over a set of terminal categories (also known as the free monoid) whenever we talk about algebras.

\begin{definition}[Heterogeneous algebras]
A \textit{heterogeneous algebra} $(T,F)$ consists of a family of sets $T=(T_i)_{i\in I}$ for some index set $I$, and a set $F$ of functions $T_{i_1}\times\dots\times T_{i_n}\to T_i$ for some indices $i_1,\dots,i_n,i\in I$. The \textit{signature} of a function $(f:T_{i_1}\times\dots\times T_{i_n}\to T_i)\in F$ is defined as $\sigma(f)=(i_1,\dots,i_n,i)$.
The signature of a constant functions is the index of their codomain. A \textit{term function} of $(T,F)$ is either an element of $F$ or a (heterogeneous) superposition of functions from $F$ and projections $\pi_k:T_{i_1},\dots,T_{i_n}\to T_{i_k}$.
\end{definition}

A classical (homogeneous) universal algebra is a special case of a heterogeneous universal algebra for a singleton index set $I$.

\begin{definition}[Function call expressions]
Let $(T,F)$ be a heterogeneous algebra and $N$ a set of variables with an associated dimensionality function $\dim:N\to \mathbb N$ . A \textit{function call expression} of dimension $m\in \mathbb N$ is either
\begin{itemize}
	\item a variable $A\in N$ with $\dim(A)=m$, or
	\item a function $f\in F$ together with a tuple of function call expressions $(e_1,\dots,e_n)$, denoted by $f[e_1,\dots,e_n]$, such that $\sigma(f)=(\dim(e_1),\dots,\dim(e_n),m)$.
\end{itemize}
In particular, all constant functions in $F$ form a function call expression together with the empty tuple. The set of all function call expressions is denoted by $CallExpr(F,N)$. The set of function call expressions that do not contain any variables (elements of $N$) is denoted by $CallExpr(F)$. The evaluation $\llbracket\cdot\rrbracket: CallExpr(F)\to T$ of a variable-free function call expression is recursively defined by $\llbracket f[e_1,\dots,e_n]\rrbracket=f(\llbracket e_1\rrbracket,\dots,\llbracket e_n\rrbracket)$.
\end{definition}

\begin{definition}
A \textit{(deterministic) Abstract Grammar} $G=(T,F,N,S,\Gamma)$ consists of a heterogeneous algebra $(T,F)$, a finite set $N$ of \textit{non-terminal categories}, a \textit{start category} $S\in N$, and a set $$\Gamma=\set{g|g:N\nrightarrow CallExpr(F,N)}$$ of partial functions, called \textit{rewrite functions}, that map non-terminal categories to function call expressions with non-terminal categories as variables.
A function call expression $e\in CallExpr(F,N)$ can be rewritten in one step into a function call expression $e'\in CallExpr(F,N)$ by the application of a rewrite function $g\in \Gamma$ to a non-terminal category $A\in N$, denoted by $e\longrightarrow_{A} e'$, if $e'$ arises from $e$ by replacing exactly one appearance of $A$ in $e$ by $g(A)$.
We write $e\longrightarrow e'$ if there exist any $g$ and $A$ such that $e\longrightarrow_{A} e'$. The transitive closure of the \textit{rewrite-in-one-step relation} $\longrightarrow$ is denoted by $\longrightarrow^*$.
The language of the grammar $G$ is the set of elements of $\bigcup_iT_i$ that can be generated from the start category $S$, that is
$$\mathcal L(G) = \set{t\in T_i | T_i \in T, \exists e\in CallExpr(F): S\longrightarrow^*e \text{ and } \llbracket e\rrbracket = t}.$$
\end{definition}

A Multiple Context-free Grammar (MCFG; \cite{sekietal91}) is an example of an Abstract Grammar under a particular algebra. Let $\Sigma$ be the set of terminal categories of a MCFG and $W = \Sigma^*$. Then $(W, \text{concat})$ is the free monoid (or word monoid) over $\Sigma$ together with the concatenation operation $\text{concat} : W \times W \to W$. The family $T = (W^n)_{n\in\mathbb{N}}$ forms a heterogeneous algebra when paired with the following operations:
\begin{itemize}
	\item the concatenation operation of the free monoid,
	\item the tuple constructors $\text{list}_n : W^n \to W^n$,
	\item the projections $\pi_k^n: W^n \to W$ ($\pi_k^k(w_1, \dots, w_k, \dots, w_n) = w_k$), and
	\item all elements of $\Sigma$ as constant functions.
\end{itemize}
We call this algebra the \emph{tuple algebra} of $W$.
Now consider an Abstract Grammar using the tuple algebra of a free monoid.
Such an Abstract Grammar is strongly equivalent to a $k$-Multiple Context-free Grammar ($k$-MCFG) if the following extra conditions hold: (i) the term functions do not copy any components of their inputs, (ii) the tuple dimension is upper-bounded by $k$, and (iii) the dimension of the start category is 1.

\begin{definition}
A \textit{probabilistic Abstract Grammar} is an \textit{Abstract Grammar} where each non-terminal category $A\in N$ is associated with a random variable $X_A$ over rewrite functions.

The probability of a function call expression $e\in CallExpr(F,N)$ rewriting in one step into a function call expression $e'\in CallExpr(F,N)$ by applying a rewrite function $g\in\Gamma$ to a non-terminal $A \in N$ is
\[
	P(e \longrightarrow_{g(A)} e') = \text{splits}(g,A,e,e') \cdot P(X_A = g),
\]
where
$\text{splits}(g,A,e,e')$ denotes the number of occurrences of the non-terminal category $A$ in the function call expression $e$ to which $g$ can be applied to get $e'$.
The probability of any function call expression $e$ rewriting into any function call expression $e'$ is
\[
	P(e \longrightarrow_{g(A)} e') = \sum_{g\in \Gamma}\sum_{A \in N} P(e \longrightarrow_A e').
\]
The probability of a function call expression $e$ rewriting into a function call expression $e'$ in any steps is
$$P(e\longrightarrow^* e'') = \sum_{e'\in CallExpr(F,N)} P(e\longrightarrow^* e')P(e'\longrightarrow e'').$$
Finally, the probability of an element $t\in T_i$ ($T_i \in T$) is
$P(t)=\sum_{\llbracket e\rrbracket = t} P(S\longrightarrow^* e)$.
\end{definition}

\subsection{Reduction automata for Abstract Grammars}
Let $G=(T,F,N,S,\Gamma)$ be an Abstract Grammar. We show in the following how to construct a reduction automaton that accepts exactly the sequences that are generatable by $G$. We choose the set of states $\mathcal S=N^*$. The transition function $\text{tran}:\mathcal S \times  N^* \nrightarrow \mathcal S$ is concatenation, just as in the context-free case. The completion function $\text{comp}:\mathcal S \to 2^{N \times \Gamma}$ is defined by
$$(A, g)\in \text{comp}(N_1,\dots,N_n)
	\quad\iff\quad
	\exists f\in F: g(A)=f[N_1,\dots,N_n] $$
for non-terminal categories $N_1,\dots,N_n$ and by
$$(A, g)\in \text{comp}(t)
	\quad\iff\quad
	\exists f\in F: g(A)=f[~] \text{ and } f()=t$$
for all singleton tuples $t\in T_1$.
Note that we assume that the leaves of a parse tree are always singleton tuples.
Any tree with a leaf that rewrites to a tuple of length higher than 1 can always be converted to a tree with only singleton leaves by adding binary branching rules using the tuple constructors.
The completion function is extended to the probabilistic case by changing its codomain to $2^{N \times \Gamma \times [0,1]}$, so that $(A, g, p) \in \text{comp}(s)$ iff the statements defined above hold and $P(X_A = g) = p$.

Remark that a reduction automaton does not define a parsing algorithm, but specifies the abstract grammar interface. To gain computational efficiency, the transition function can, just as in the context-free case, rigorously filter so that it only yields states that eventually lead to completions.
The abstract grammar interface derived from this automata is the same as in the context-free case.

\subsection{Parsing Abstract Grammars}
To construct a parsing algorithm for Abstract Grammars, we have to know how the term functions act on the indices of the input sequence. For the tuple algebra $T=(W^n)_{n\in\mathbb N}$, each tuple of words is potentially associated with a tuple of pairs of natural numbers describing the start and the end index for every word in the tuple.

More formally, given an input word $w\in W$ of length $m=|w|$, we define a partial (heterogeneous) algebra $R$, called the \textit{range algebra} of length $m$, and a homomorphism $\varrho:R\to T$ that models the indexing as follows. Denote the less-then relation on the set $\set{1,\dots,m+1}$ by $L$. Thus, $(i,j)\in L$ iff $0<i<j\leq m+1$ for natural numbers $i$ and $j$. We set $R=(L^n)_{n\in\mathbb N}$.
It is clear how to define the tuple constructors $\text{list}_n:L^n\to L^n$, the projections $\pi^n_k:L^n\to L$, and all constants. The concatenation on the range algebra is a partial function defined by $\text{concat}((i,j),(k,l))=(i,l)$ iff $j=k$. The homomorphism $\varrho$ on the set $L$ is given by $\varrho((i,j)) = w_{(i,j)}$ where $w_{(i,j)}$ denotes the subsequence of $w$ beginning on the $i$-th terminal category and ending on the $j-1$-th category.
In particular, $w_{(1,m+1)}=w$ and $w_{(i,i+1)}$ is the $i$-th terminal symbol of $w$. $\varrho$ extends naturally from $L$ to $L^n$, since it must commute over the tuple constructors.
Remark that $\varrho$ is in 1-to-1 relation to $w$.

The parsing as deduction rules are shown in figure \ref{fig:abstract.deduction}. In contrast to the context-free case, we equip edges and constituents with ranges and constrain the edge completion on the definiteness of the respective term function.

\begin{figure}[!htb]
\begin{alignat*}{3}
&\text{items:} && \text{edges}\qquad\qquad[s,r_1,\dots,r_n] \quad\,\text{ for } s\in \mathcal S \text{ and } r_i\in R \\
& &&\text{constituents}\quad[A,r] \quad\quad\qquad~~\text{ for } A\in N \text{ and } r\in R \\ \\
&\text{goal item:} && [S,(1,|w|+1)] \\ \\
&\text{axioms:} && \frac{}{[w_i,(i,i+1)]}  \quad\text{ for } i\in\{1,\dots,|w|\} \\ \\
&\text{introduce edge:} && \frac{[A,r]}{[s,r]} \quad s=\text{tran}(s_0, A) \\ \\
&\text{complete edge:} && \frac{[s,r_1,\dots,r_n]}{[A,f(r_1,\dots,r_n)]} \quad (A, g) \in \text{comp}(s), g(A) = f[s], \text{ and $f$ is defined on $(r_1, \dots, r_n)$} \\ \\
&\text{fundamental rule:}\qquad\qquad && \frac{[s,r_1,\dots,r_n] \quad [A,r]}{[s',r_1,\dots,r_n,r]} \quad \text{tran}(s,A) = s'
\end{alignat*}
\caption{Description of the parsing algorithm by the use of deduction rules}
\label{fig:abstract.deduction}
\end{figure}

\subsection{Implementation}

The implementation given in as \S\ref{sec:implementation} and figure 4 is generally sufficient to be applied to the fully general Abstract Grammars.
Only two types, \texttt{EdgeKey} and \texttt{ConstituentKey} must be altered.
For context-free parsing, these two types have attributes \texttt{start} and \texttt{end\_} to keep track of the substring of the input sequence they are associated with.
For the general case, we must alter these types so that they keep track of a vector of start and end indices.

\vspace{10px}
\begin{lstlisting}
immutable EdgeKey{St}
    state :: St
    ranges :: Vector{Vector{Tuple{Int, Int}}}
end
immutable ConsKey{C}
    cat   :: C
    ranges :: Vector{Tuple{Int, Int}}
end
\end{lstlisting}

\section{Minimalist Grammar Interface}
\label{sec:mgs}
In this section, we give an example of how to write an interface for a specific grammar formalism. The formalism we use is the Minimalist Grammar (MG), introduced by \citealt{stable97}.\footnote{See also \citealt{stable11} for a good introduction to this formalism.} MGs are weakly equivalent to MCFGs \citep{michae98}. In the following, we define MGs and show how to formulate the correspondence between an MG and the abstract grammar interface.\footnote{The definition given here is taken from \citealt{harkem01diss}, with one alteration. We distinguish between two kinds of selectors (left-selectors $\overset{L}{=}f$ and right-selectors $\overset{R}{=}f$). We use this because it allows for a more straightforward analysis of linguistic phenomena with complements on the left or right, and corresponds to the use of back- and forward-slashes in categorial grammar.} We subsequently give an implementation of the interface in Julia.

\begin{definition}
    A Minimalist Grammar is a pair $G = \langle LEX, \{\mathbf{merge}, \mathbf{move}\}\rangle$. $LEX \subseteq \Sigma \times \mathcal F^*$ is a lexicon where:
    \begin{itemize}
      \item $\Sigma$ is a vocabulary,
      \item $\mathcal{F}$ is a set of syntactic features consisting of:
      \begin{itemize}
        \item Selectees of the form \texttt{f} (for any symbol \texttt{f}),
        \item Selectors of the form \texttt{f=} and \texttt{=f},
        \item Licensors of the form \texttt{+f}, and
        \item Licensees of the form \texttt{-f}.
      \end{itemize}
    \end{itemize}
    We denote a lexical item $(\alpha, \beta) \in LEX$ by $\alpha: \beta$.
    $\mathbf{merge}$ and $\mathbf{move}$ are structure building operations defined as follows using natural deduction notation. Let $C = \Sigma^* \times \mathcal{F}^*$ be the set of possible \emph{chain} and $E = C^*$ be the set of possible \emph{expression}.
    \begin{itemize}
      \item $\mathbf{merge}: (E \times E) \to E$ is the union of the following three functions. For any $\gamma \in \mathcal{F}^*, \delta \in \mathcal{F}^+$ and any chains $\alpha_1, \dots,  \alpha_k, \iota_1, \dots, \iota_l$ ($0 \leq k, l$):
      \[
      \mathbf{merge}_{R1}: \quad
      \frac{s :\ \texttt{=f} \gamma, \alpha_1, \dots, \alpha_k \quad t : \texttt{f}, \iota_1, \dots, \iota_l}{st : \gamma, \alpha_1, \dots, \alpha_k,  \iota_1, \dots, \iota_l}
      \]
      \[
      \mathbf{merge}_{L1}: \quad
      \frac{s :\ \texttt{f=} \gamma, \alpha_1, \dots, \alpha_k \quad t : \texttt{f}, \iota_1, \dots, \iota_l}{ts : \gamma, \alpha_1, \dots, \alpha_k, \iota, \dots, \iota_l}
      \]
      \[
      \mathbf{merge}_{R2}: \quad
      \frac{s :\ \texttt{=f} \gamma, \alpha_1, \dots, \alpha_k \quad t : \texttt{f} \delta, \iota_1, \dots, \iota_l}{s : \gamma, \alpha_1, \dots, \alpha_k, t :\delta, \iota, \dots, \iota_l}
      \]
      \[
      \mathbf{merge}_{L2}: \quad
      \frac{s :\ \texttt{f=} \gamma, \alpha_1, \dots, \alpha_k \quad t : \texttt{f} \delta, \iota_1, \dots, \iota_l}{s : \gamma, \alpha_1, \dots, \alpha_k, t :\delta, \iota, \dots, \iota_l}
      \]
      \item $\mathbf{move}: E \to E$ is the union of the following two functions. For any $\gamma \in \mathcal{F}^*, \delta \in \mathcal{F}^+$ and any chains $\alpha_1, \dots, \alpha_k$ satisfying the Shortest Move Constraint
      \begin{equation}
        \text{None of the chains $\alpha_i$ has $-f$ as its first feature}
        \tag{\text{SMC}}
      \end{equation}
      \[
      \mathbf{move}_1: \quad
      \frac{s : \texttt{+f}\gamma, \alpha_1, \dots, \alpha_{i-1}, t : \texttt{-f}, \alpha_{i+1}, \dots, \alpha_k}{ts: \gamma, \alpha_1, \dots, \alpha_{i-1},\alpha_{i+1}, \dots, \alpha_k}
      \]
      \[
      \mathbf{move}_2: \quad
      \frac{s : \texttt{+f}\gamma, \alpha_1, \dots, \alpha_{i-1}, t : \texttt{-f}\delta, \alpha_{i+1}, \dots, \alpha_k}{s : \gamma, \alpha_1, \dots, \alpha_{i-1}, t : \delta, \alpha_{i+1}, \dots, \alpha_k}
      \]
    \end{itemize}
\end{definition}

A derivation in a Minimalist Grammar is centered around feature-checking.
Checking a feature sequence $\texttt{f}_1\dots \texttt{f}_k$ against another feature sequence $\texttt{g}_1\dots \texttt{g}_l$ corresponds to verifying whether $\texttt{f}_1$ and $\texttt{g}_1$ match and, if so, deleting them, and returning the tail of each sequence.
The \textbf{merge} operation takes two sequences of chains, checks the respective feature sequences of their head chain, and potentially concatenates the respective vocabulary sequences of their head chain.
An MG with only the \textbf{merge} operation is weakly equivalent to a CFG.
The \textbf{move} operation takes one chain sequence, who's head chain's head feature is a licensor feature, checks whether it contains a chain whose head feature is a matching licensee feature, and potentially concatenates the two chains' respective vocabulary items in reverse order.
This operation is what allows MGs to capture non-local dependencies, making them more powerful than CFGs.

\begin{figure}
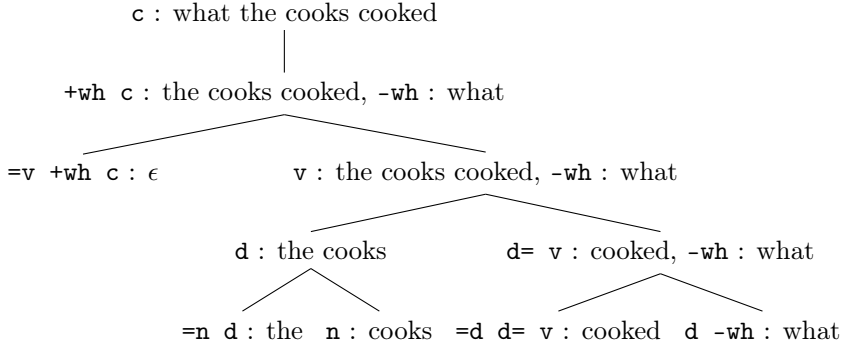

\centering
  \Tree [.{\texttt{c} : what the cooks cooked} [.{\texttt{+wh c} : the cooks cooked, \texttt{-wh} : what}
          [.{\texttt{=v +wh c} : $\epsilon$} ] [.{\texttt{v} : the cooks cooked, \texttt{-wh} : what}
            [.{\texttt{d} : the cooks} [.{\texttt{=n d} : the} ] [.{\texttt{n} : cooks} ] ]
            [.{\texttt{d= v} : cooked, \texttt{-wh} : what}
              [.{\texttt{=d d= v} : cooked} ] [.{\texttt{d -wh} : what} ]
            ]
          ]
        ] ]

  \caption{MG derivation for `what the cooks cooked' such that `what' moves from the complement position of `cooked', The grammar is such that $LEX$ =
    \{cooked : \texttt{=d d= v},
    who : \texttt{d -wh},
    cooks : \texttt{n},
    the : \texttt{=n d},
    $\epsilon$ : \texttt{=v +wh c}\}
  }
  \label{fig:mgderivation}
\end{figure}

An MG \emph{derivation tree} represents a record of the structure building operations performed in a particular derivation (or parse) of a sentence. An example derivation  tree is given in figure \ref{fig:mgderivation}.

\begin{definition}
		A \emph{probabilistic Minimalist Grammar} is a triple $\langle LEX, \{\textbf{merge}, \textbf{move}\}, P\rangle$ consisting of a Minimalist Grammar and a probability distribution over lexical items $P: LEX \to [0,1]$.
		The probability of the result of a merge operation is the product of the input probabilities:
		\[ P(\textbf{merge}(\alpha, \beta)) = P(\alpha)P(\beta). \]
		The probability of the result of a move operation is the probability of its input:
		\[ P(\textbf{move}(\alpha)) = P(\alpha). \]
\end{definition}

\subsection{Abstract interface}
\label{sec:interface}

Our reduction automaton makes reference to the operations in a tuple algebra.
So in order to specify interface functions for our Minimalist Grammar, we must specify the tuple operations that correspond to the MG structure-building operations.
Figure \ref{fig:mgfunctions} builds the set $F$ of operations.
We must also assume we are given a pairing between non-constant tuple operations and their associated structure building operation.
Let $p$ be a relation denoting this pairing, so that $p(f, o)$ denotes that $f \in F$ is paired with $o \in \{\mathbf{merge}_{R1}, \mathbf{merge}_{R2}, \mathbf{merge}_{L1}, \mathbf{merge}_{L2}, \mathbf{move}_1, \mathbf{move}_2\}$.

\begin{figure}
  Let $(T, F)$ be a tuple algebra where $T = \{T_1, \dots, T_k\}$ such that $T_i \subseteq \Sigma^i$.
	Let $F$ consist of the following operations.
	For each lexical item $\alpha:\beta \in LEX$, define the constant function $f_{\alpha:\beta}() = (\alpha)$.
  For each $( s_1, \dots, s_m ), ( t_1, \dots, t_m ),  \in \bigcup T_i$ ($1 \leq m \leq k$) define the following operations corresponding to the structure building operations:
  \begin{align*}
    ( s_1, \dots, s_m ), ( t_1, \dots, t_m ) &\mapsto ( s_1t_1, s_2, \dots, s_m, t_2,  \dots, t_m) && (\mathbf{merge}_{R1}) \\
    ( s_1, \dots, s_m ), ( t_1, \dots, t_m ) &\mapsto ( t_1s_1, s_2, \dots, s_m, t_2,  \dots, t_m) && (\mathbf{merge}_{L1}) \\
    ( s_1, \dots, s_m ), ( t_1, \dots, t_m ) &\mapsto ( s_1, \dots, s_m, t_1, \dots, t_m ) && (\mathbf{merge}_{R2/L2}) \\
    ( s_1, \dots, s_m ) &\mapsto ( s_1s_i, s_2, \dots, s_{i-1},  s_{i+1}, \dots,  s_m) \quad (1 < i \leq m) &&  (\mathbf{move}_1) \\
    ( s_1, \dots, s_m ) &\mapsto ( s_1, \dots, s_m) &&  (\mathbf{move}_2)
  \end{align*}
  \caption{tuple operations for each lexical item and structure building operation in an MG.}
  \label{fig:mgfunctions}
\end{figure}

We must specify five interface functions: tran\_possible, tran, comp, startstates, startcategories.
The constant function startcategories is user defined, though the traditional choice for a start category in an MG is \texttt{c}.
Recall that a state for an Abstract Grammar is a sequence of non-terminals.
Abstract nonterminals correspond to tuples of feature sequences in an MG $(F^*)^*$, so the set of states is
\[ \mathcal{S} \subseteq ((F^*)^*)^*. \]
There is one start state, the empty sequence:
\[ \text{startstates}() = \{\epsilon\}. \]

Now we turn to the transition and completion functions.
A transition is meant to move the parser one step through a rule.
In an MG, this corresponds to moving from left to right through the top portion of a \textbf{merge} deduction rule. The tran\_possible function is true iff it is transitioning from the start state (and has thus encountered a possible left hand side of a merge rule), or, if not, has encountered a matchinig right hand side of a merge rule.
\begin{equation*}
  \text{tran\_possible}(s, \gamma) \iff
    s \in \text{startstates}() \text{ or }
    \mathbf{merge}(s, \gamma) \text{ is defined}
\end{equation*}
The tran function concatenates sequences of features assuming that tran\_possible was true.\footnote{Some notation: let $\circ$ signify concatenation of an element onto the end of a vector. So that, for example: $\langle a, b, c\rangle \circ d = \langle a, b, c, d\rangle$}.
If $s$ is the start state, then $s = \epsilon$, so tran gives a singleton sequence.
Otherwise, tran gives a pair.
These are the only two cases, since MG trees are maximally binary branching.
\begin{equation*}
  \text{tran}(s, \gamma) =
    s \circ \gamma
\end{equation*}

A completion is meant to correspond to a constituent in the parsed structure, thus each node in the MG derivation tree corresponds to the output of a completion.
Therefore, a completion occurs whenever any of the structure building operations apply or a lexical item is encountered.
For any $w \in \Sigma$,
\begin{equation*}
  \text{comp}(\langle w \rangle) = \{(\alpha, \alpha \mapsto f_{w:\alpha}[]) | (w, \alpha) \in LEX\}
\end{equation*}
For any sequences of features $\alpha, \gamma, \delta \in F^+$,
\begin{align*}
  (\alpha, \alpha \mapsto f[\gamma, \delta]) \in \text{comp}(\langle \gamma, \delta \rangle) &\iff \exists f \in F~\exists i \in \mathbb{N}: p(f, \mathbf{merge}_i) \text{ and } \mathbf{merge}_i(\gamma, \delta) = \alpha,\\
  (\alpha, \alpha \mapsto f[\gamma]) \in \text{comp}(\langle \gamma \rangle) &\iff \exists f \in F~\exists i \in \mathbb{N}: p(f, \mathbf{move}_i) \text{ and } \mathbf{move}_i(\gamma) = \alpha.
\end{align*}

\subsection{Julia implementation}
\label{sec:julia}

A Minimalist Grammar is implemented as a parametric type \texttt{MinimalistGrammar\{T, Score\}} where \texttt{T} is the type of the vocabulary items and \texttt{Score} is the type of the score associated with a lexical item (e.g., a probability).
Typically, \texttt{T} is set to \texttt{String} and \texttt{Score} is set to some numerical type.
It relies on a type \texttt{LexicalItem} and a type \texttt{Feature}.
A \texttt{Feature} may be any arbitrary data type which includes the functions defined on it below; we let features be strings.

\vspace{10px}
\begin{lstlisting}
Feature = String # features are strings of the form =f, f=, +f, -f, f
selects_right(f :: Feature) = length(f) > 2 && f[1] == '='
selects_left(f :: Feature) = length(f) > 2 && last(f) == '='
is_selector(f :: Feature) = selects_right(f) || selects_left(f)
is_licensor(f :: Feature) = length(f) > 1 && f[1] == '+'
is_licensee(f :: Feature) = length(f) > 1 && f[1] == '-'
is_selectee(f :: Feature) = !is_selector(f) && !is_licensor(f) && !is_licensee(f)
name(f :: Feature) = if is_selector(f)
                        f[3:end]
                     elseif is_licensor(f) || is_licensee(f)
                        f[2:end]
                     else f[1:end]
                     end

type LexicalItem{T, Score}
  phon :: T
  features :: Vector{Feature}
  score :: Score
end

type MinimalistGrammar{T, Score}
  lexicon :: Vector{LexicalItem{T, Score}}
  start_symbols :: Vector{Vector{Feature}}
  tuple_operations :: Vector{TupleOperation}
end
\end{lstlisting}
\vspace{10px}

We also define functions associated with the structure building operations \textbf{merge} and \textbf{move}. In the interest of space we include just the type signatures for these functions here.

\vspace{10px}
\begin{lstlisting}
function merge(f :: Vector{Vector{Feature}}, g :: Vector{Vector{Feature}})
function move(f :: Vector{Vector{Feature}})
\end{lstlisting}

The type \texttt{MinimalistState} implements a state for a Minimalist Grammar, which consists of a sequence of tuples of feature sequences $((\mathcal{F}^*)^*)^*$.

\vspace{10px}
\begin{lstlisting}
type MinimalistState
  categories :: Vector{Vector{Vector{Feature}}}
  isfinal :: Bool
end
\end{lstlisting}
\vspace{10px}

Now we can give the implementations of the interface functions. The implementations for startsymbols and startstates, are trivial.

\vspace{10px}
\begin{lstlisting}
startsymbols{T,Sc}(g :: MinimalistGrammar{T,Sc}) = map(x -> [x], start_symbols(g))
startstate{T,Sc}(g :: MinimalistGrammar{T,Sc}) = MinimalistState([], false)
\end{lstlisting}
\vspace{10px}

The implementations for tran (\texttt{transition}) and tran\_possible (\texttt{is\_possible\_transition}) are given below.
They rely on function
\texttt{match(f :: Vector\{Feature\}, g :: Vector\{Feature\})}, which is true iff the two features match.

\vspace{10px}
\begin{lstlisting}
function is_possible_transition{T,Sc}(g :: MinimalistGrammar{T,Sc},
                                      s :: MinimalistState,
                                      c :: Vector{Vector{Feature}})
  isempty(categories(s)) || match(last(categories(s))[1], c[1])
end

function transition{T,Sc}(g :: MinimalistGrammar{T,Sc}, s :: MinimalistState,
                          c :: Vector{Vector{Feature}})
  if isempty(categories(s))
    MinimalistState([c], false)
  else
    MinimalistState([last(categories(s)), c], true)
  end
end
\end{lstlisting}
\vspace{10px}

The completion function is the most involved and is separated into two definitions, one for states and the other for lexical items.
The return type of the \texttt{completions} function is a vector of triples the first entry is a tuple of feature sequences, the second represents the rule associated with the completion (here represented whose first entry is the category of the bottom portion of the deduction rule, and whose second entry is the sequence of categories of the top portion of the rule), and the third is the score of the rule.
These function implementations rely on the following functions:
\begin{itemize}
  \item \texttt{is\_movable(f :: Vector\{Vector\{Feature\}\})}: true iff the head of \texttt{f} is a licensor feature and \texttt{f} contains a matching licensee feature.
  \item \texttt{args(o :: TupleOperation)}: returns the number of arguments of the tuple operation.
  \item \texttt{dims(o :: TupleOperation)}: returns a vector of the dimensions of the operations arguments.
  \item \texttt{is\_lexical(T :: DataType, o :: TupleOperation)}: true iff \texttt{o} is a constant function which returns a vocabulary item of type \texttt{T}.
\end{itemize}

\newpage
Here is the implementation for states:
\vspace{10px}
\begin{lstlisting}
function completions{T,Sc}(g :: MinimalistGrammar{T,Sc}, s :: MinimalistState)
  # initialize return value
  C = Vector{Vector{Feature}}
  R = Tuple{C, TupleOperation, Vector{C}}
  ret = Vector{Tuple{C, R, score_type(g)}}()

  # make a completion if you can apply merge
  for o in tuple_operations(g)
    if length(categories(s)) == 2 &&
        match(categories(s)[1][1], categories(s)[2][1]) &&
        is_correct_tupleoperation(categories(s)[1], categories(s)[2], o)

      merged = merge(categories(s)...)
      if length(output(o)) == length(merged)
        push!(ret, (merged, (merged, o, categories(s)), score_type(g)(1)))
      end
    end
  end

  # make a completion if you can apply move
  for o in tuple_operations(g)
    if length(categories(s)) == 1 &&
          is_movable(categories(s)[1]) &&
          args(o) == 1 &&
          dims(o)[1] == length(categories(s)[1]) &&
          !is_lexical(terminal_type(g), o)

      moved = move(categories(s)[1])
      if (length(moved) == length(categories(s)[1]) && o == mg_move_nonfinal) ||
          (length(moved) != length(categories(s)[1]) && o == mg_move_final

        push!(ret, (moved, (moved, o, categories(s)), score_type(g)(1)))
      end
    end
  end

  return ret
end
\end{lstlisting}
\vspace{10px}
And here is the implementation for lexical items:
\vspace{10px}
\begin{lstlisting}
function completions{T,Sc}(g :: MinimalistGrammar{T,Sc}, word :: T)
  C = Vector{Vector{Feature}}
  R = Tuple{C, Vector{C}}
  ret = Vector{Tuple{C, R, score_type(g)}}()

  is_term_op(o::TupleOperation) = (
    is_lexical(terminal_type(g), o) &&
    output(o) == [[word]])
  for o in filter(is_term_op, tuple_operations(g))
    for l in lexicon(g)
      if phon(l) == word
        push!(ret, ([features(l)], ([features(l)], o, [[features(l)]]), score(l)))
      end
    end
  end
  return ret
end
\end{lstlisting}

\section{Conclusion}

In this technical report we presented a general framework for parsing consisting of (i) the Abstract Grammar formalism, capable of expressing a variety of formal grammars up to at least MG-equivalent grammars, (ii) a reduction automaton for Abstract Grammars, (iii) an abstract grammar interface derived from the reduction automaton, suitable for many different grammars, and (iv) our Julia implementation of the parser.
We also gave an example of an interface for a Minimalist Grammar, presenting both the abstract interface for the formal definition of Minimalist Grammars and its implementation in Julia.

\bibliography{../everything.bib}

\begin{thebibliography}{}

\bibitem [\protect \citeauthoryear {%
Bezanson%
, Edelman%
, Karpinski%
\BCBL {}\ \BBA {} Shah%
}{%
Bezanson%
\ \protect \BOthers {.}}{%
{\protect \APACyear {2017}}%
}]{%
bezanson2017julia}
\APACinsertmetastar {%
bezanson2017julia}%
\begin{APACrefauthors}%
Bezanson, J.%
, Edelman, A.%
, Karpinski, S.%
\BCBL {}\ \BBA {} Shah, V\BPBI B.%
\end{APACrefauthors}%
\unskip\
\newblock
\APACrefYearMonthDay{2017}{}{}.
\newblock
{\BBOQ}\APACrefatitle {Julia: A fresh approach to numerical computing} {Julia:
  A fresh approach to numerical computing}.{\BBCQ}
\newblock
\APACjournalVolNumPages{SIAM Review}{59}{1}{65--98}.
\PrintBackRefs{\CurrentBib}

\bibitem [\protect \citeauthoryear {%
Goodman%
}{%
Goodman%
}{%
{\protect \APACyear {1998}}%
}]{%
Goodman1998}
\APACinsertmetastar {%
Goodman1998}%
\begin{APACrefauthors}%
Goodman, J\BPBI T.%
\end{APACrefauthors}%
\unskip\
\newblock
\APACrefYear{1998}.
\unskip\
\newblock
\APACrefbtitle {{Parsing inside-out}} {{Parsing inside-out}}\
  \APACtypeAddressSchool {\BPhD}{}{Harvard University}.
\unskip\
\newblock
\begin{APACrefURL} \url{https://arxiv.org/abs/cmp-lg/9805007} \end{APACrefURL}
\PrintBackRefs{\CurrentBib}

\bibitem [\protect \citeauthoryear {%
Harkema%
}{%
Harkema%
}{%
{\protect \APACyear {2001}}%
}]{%
harkem01diss}
\APACinsertmetastar {%
harkem01diss}%
\begin{APACrefauthors}%
Harkema, H.%
\end{APACrefauthors}%
\unskip\
\newblock
\APACrefYear{2001}.
\unskip\
\newblock
\APACrefbtitle {Parsing Minimalist Languages} {Parsing minimalist languages}\
  \APACtypeAddressSchool {\BUPhD}{}{}.
\unskip\
\newblock
\APACaddressSchool {}{University of California, Los Angeles}.
\PrintBackRefs{\CurrentBib}

\bibitem [\protect \citeauthoryear {%
Joshi%
}{%
Joshi%
}{%
{\protect \APACyear {1985}}%
}]{%
joshi85}
\APACinsertmetastar {%
joshi85}%
\begin{APACrefauthors}%
Joshi, A\BPBI K.%
\end{APACrefauthors}%
\unskip\
\newblock
\APACrefYearMonthDay{1985}{}{}.
\newblock
{\BBOQ}\APACrefatitle {Tree adjoining grammars: How much context-sensitivity is
  required to provide reasonable structural descriptions?} {Tree adjoining
  grammars: How much context-sensitivity is required to provide reasonable
  structural descriptions?}{\BBCQ}
\newblock
\BIn{} D\BPBI R.~Dowty, L.~Karttunen\BCBL {}\ \BBA {} A\BPBI M.~Zwicky\
  (\BEDS), \APACrefbtitle {Natural Language Parsing.} {Natural language
  parsing.}
\newblock
\APACaddressPublisher{}{Cambridge University Press}.
\PrintBackRefs{\CurrentBib}

\bibitem [\protect \citeauthoryear {%
Klein%
\ \BBA {} Manning%
}{%
Klein%
\ \BBA {} Manning%
}{%
{\protect \APACyear {2004}}%
}]{%
klein2004parsing}
\APACinsertmetastar {%
klein2004parsing}%
\begin{APACrefauthors}%
Klein, D.%
\BCBT {}\ \BBA {} Manning, C\BPBI D.%
\end{APACrefauthors}%
\unskip\
\newblock
\APACrefYearMonthDay{2004}{}{}.
\newblock
{\BBOQ}\APACrefatitle {Parsing and hypergraphs} {Parsing and
  hypergraphs}.{\BBCQ}
\newblock
\APACjournalVolNumPages{New developments in parsing technology}{}{}{351--372}.
\PrintBackRefs{\CurrentBib}

\bibitem [\protect \citeauthoryear {%
Lerdahl%
\ \BBA {} Jackendoff%
}{%
Lerdahl%
\ \BBA {} Jackendoff%
}{%
{\protect \APACyear {1985}}%
}]{%
lerdahl1985generative}
\APACinsertmetastar {%
lerdahl1985generative}%
\begin{APACrefauthors}%
Lerdahl, F.%
\BCBT {}\ \BBA {} Jackendoff, R.%
\end{APACrefauthors}%
\unskip\
\newblock
\APACrefYear{1985}.
\newblock
\APACrefbtitle {A generative theory of tonal music} {A generative theory of
  tonal music}.
\newblock
\APACaddressPublisher{}{MIT press}.
\PrintBackRefs{\CurrentBib}

\bibitem [\protect \citeauthoryear {%
Michaelis%
}{%
Michaelis%
}{%
{\protect \APACyear {1998}}%
}]{%
michae98}
\APACinsertmetastar {%
michae98}%
\begin{APACrefauthors}%
Michaelis, J.%
\end{APACrefauthors}%
\unskip\
\newblock
\APACrefYearMonthDay{1998}{}{}.
\newblock
{\BBOQ}\APACrefatitle {Derivational minimalism is mildly context-sensitive}
  {Derivational minimalism is mildly context-sensitive}.{\BBCQ}
\newblock
\BIn{} \APACrefbtitle {LACL} {Lacl}\ (\BVOL~98, \BPGS\ 179--198).
\PrintBackRefs{\CurrentBib}

\bibitem [\protect \citeauthoryear {%
Pollard%
}{%
Pollard%
}{%
{\protect \APACyear {1984}}%
}]{%
pollar84}
\APACinsertmetastar {%
pollar84}%
\begin{APACrefauthors}%
Pollard, C\BPBI J.%
\end{APACrefauthors}%
\unskip\
\newblock
\APACrefYear{1984}.
\unskip\
\newblock
\APACrefbtitle {Generalized Phrase Structure Grammars, Head Grammars, and
  Natural Language} {Generalized phrase structure grammars, head grammars, and
  natural language}\ \APACtypeAddressSchool {\BUPhD}{}{}.
\unskip\
\newblock
\APACaddressSchool {}{Stanford University}.
\PrintBackRefs{\CurrentBib}

\bibitem [\protect \citeauthoryear {%
Rohrmeier%
}{%
Rohrmeier%
}{%
{\protect \APACyear {2011}}%
}]{%
rohrmeier2011towards}
\APACinsertmetastar {%
rohrmeier2011towards}%
\begin{APACrefauthors}%
Rohrmeier, M.%
\end{APACrefauthors}%
\unskip\
\newblock
\APACrefYearMonthDay{2011}{}{}.
\newblock
{\BBOQ}\APACrefatitle {Towards a generative syntax of tonal harmony} {Towards a
  generative syntax of tonal harmony}.{\BBCQ}
\newblock
\APACjournalVolNumPages{Journal of Mathematics and Music}{5}{1}{35--53}.
\PrintBackRefs{\CurrentBib}

\bibitem [\protect \citeauthoryear {%
Rohrmeier%
\ \BBA {} Neuwirth%
}{%
Rohrmeier%
\ \BBA {} Neuwirth%
}{%
{\protect \APACyear {2015}}%
}]{%
rohrmeier2015towards}
\APACinsertmetastar {%
rohrmeier2015towards}%
\begin{APACrefauthors}%
Rohrmeier, M.%
\BCBT {}\ \BBA {} Neuwirth, M.%
\end{APACrefauthors}%
\unskip\
\newblock
\APACrefYearMonthDay{2015}{}{}.
\newblock
{\BBOQ}\APACrefatitle {Towards a Syntax of the Classical Cadence} {Towards a
  syntax of the classical cadence}.{\BBCQ}
\newblock
\BIn{} M.~Neuwirth\ \BBA {} P.~Berg\'e\ (\BEDS), \APACrefbtitle {What is a
  Cadence?} {What is a cadence?}
\newblock
\APACaddressPublisher{}{Leuven University Press}.
\PrintBackRefs{\CurrentBib}

\bibitem [\protect \citeauthoryear {%
Seki%
, Matsumura%
, Fujii%
\BCBL {}\ \BBA {} Kasami%
}{%
Seki%
\ \protect \BOthers {.}}{%
{\protect \APACyear {1991}}%
}]{%
sekietal91}
\APACinsertmetastar {%
sekietal91}%
\begin{APACrefauthors}%
Seki, H.%
, Matsumura, T.%
, Fujii, M.%
\BCBL {}\ \BBA {} Kasami, T.%
\end{APACrefauthors}%
\unskip\
\newblock
\APACrefYearMonthDay{1991}{}{}.
\newblock
{\BBOQ}\APACrefatitle {On multiple context-free grammars} {On multiple
  context-free grammars}.{\BBCQ}
\newblock
\APACjournalVolNumPages{Theoretical Computer Science}{88}{2}{191--229}.
\PrintBackRefs{\CurrentBib}

\bibitem [\protect \citeauthoryear {%
Shieber%
}{%
Shieber%
}{%
{\protect \APACyear {1985}}%
}]{%
shiebe85}
\APACinsertmetastar {%
shiebe85}%
\begin{APACrefauthors}%
Shieber, S\BPBI M.%
\end{APACrefauthors}%
\unskip\
\newblock
\APACrefYearMonthDay{1985}{}{}.
\newblock
{\BBOQ}\APACrefatitle {Evidence against the context-freeness of natural
  language} {Evidence against the context-freeness of natural language}.{\BBCQ}
\newblock
\APACjournalVolNumPages{The Formal complexity of natural
  language}{33}{}{320--332}.
\PrintBackRefs{\CurrentBib}

\bibitem [\protect \citeauthoryear {%
Shieber%
, Schabes%
\BCBL {}\ \BBA {} Pereira%
}{%
Shieber%
\ \protect \BOthers {.}}{%
{\protect \APACyear {1993}}%
}]{%
Shieber1993}
\APACinsertmetastar {%
Shieber1993}%
\begin{APACrefauthors}%
Shieber, S\BPBI M.%
, Schabes, Y.%
\BCBL {}\ \BBA {} Pereira, F\BPBI C.%
\end{APACrefauthors}%
\unskip\
\newblock
\APACrefYearMonthDay{1993}{}{}.
\newblock
{\BBOQ}\APACrefatitle {{Principles and Implementation of Deductive Parsing}}
  {{Principles and Implementation of Deductive Parsing}}.{\BBCQ}
\newblock
\APACjournalVolNumPages{Journal of Logic Programming}{}{}{}.
\PrintBackRefs{\CurrentBib}

\bibitem [\protect \citeauthoryear {%
Stabler%
}{%
Stabler%
}{%
{\protect \APACyear {1997}}%
}]{%
stable97}
\APACinsertmetastar {%
stable97}%
\begin{APACrefauthors}%
Stabler, E.%
\end{APACrefauthors}%
\unskip\
\newblock
\APACrefYearMonthDay{1997}{}{}.
\newblock
{\BBOQ}\APACrefatitle {Derivational minimalism} {Derivational
  minimalism}.{\BBCQ}
\newblock
\BIn{} C.~Retor\'e\ (\BED), \APACrefbtitle {Logical Aspects of Computational
  Linguistics} {Logical aspects of computational linguistics}\ (\BVOL\ LNCS
  No.~1328, \BPGS\ 68--95).
\newblock
\APACaddressPublisher{Berlin}{Springer}.
\PrintBackRefs{\CurrentBib}

\bibitem [\protect \citeauthoryear {%
Stabler%
}{%
Stabler%
}{%
{\protect \APACyear {2011}}%
}]{%
stable11}
\APACinsertmetastar {%
stable11}%
\begin{APACrefauthors}%
Stabler, E.%
\end{APACrefauthors}%
\unskip\
\newblock
\APACrefYearMonthDay{2011}{}{}.
\newblock
{\BBOQ}\APACrefatitle {Computational perspectives on minimalism} {Computational
  perspectives on minimalism}.{\BBCQ}
\newblock
\APACjournalVolNumPages{Oxford handbook of linguistic
  minimalism}{}{}{617--643}.
\PrintBackRefs{\CurrentBib}

\end{thebibliography}
\bibliographystyle{apacite}

\end{document}